\documentclass[11pt]{article}

\usepackage[preprint]{acl}

\usepackage{times}
\usepackage{latexsym}
\usepackage{booktabs}

\usepackage{listings}
\usepackage[T1]{fontenc}
\usepackage[utf8]{inputenc}

\usepackage{microtype}

\usepackage{inconsolata}

\usepackage{graphicx}

\usepackage[tickmarkheight=2pt,colorinlistoftodos]{todonotes}
\setuptodonotes{inline,color=yellow!40}
\title{ACL-Verbatim: hallucination-free question answering for research}

\author{
  \textbf{Gábor Recski\textsuperscript{1,2}},
  \textbf{Szilveszter Tóth\textsuperscript{2}},
  \textbf{Nadia Verdha\textsuperscript{1}},
  \textbf{István Boros\textsuperscript{2}},
  \textbf{Ádám Kovács\textsuperscript{2}}
\\
  \textsuperscript{1}TU Wien,
  \textsuperscript{2}KR Labs
\\
\small{
    \textbf{Correspondence:} \href{mailto:recski@krlabs.eu}{recski@krlabs.eu}
}
}

\begin{document}
\maketitle
\begin{abstract}
Academic researchers need efficient and reliable methods for collecting high-quality information from trusted sources, but modern tools for AI-assisted research still suffer from the tendency of Large Language Models (LLMs) to produce factually inaccurate or nonsensical output, commonly referred to as hallucinations. We apply the extractive question answering system VerbatimRAG \citep{Kovacs:2025b} to research papers in the ACL Anthology\footnote{\url{https://aclanthology.org/}}, directly mapping user queries to verbatim text spans in retrieved documents. We contribute a novel ground truth dataset for the task of mapping user queries to relevant text spans in research papers, and use it to train and evaluate a variety of extractive models. Human annotation is performed by NLP researchers and is based on synthetic user queries generated using a custom pipeline based on the ScIRGen methodology \citep{Lin:2025}, paired with chunks of research papers retrieved by VerbatimRAG. On this benchmark, a $150$M-parameter ModernBERT token classifier trained on silver supervision from our pipeline achieves the best word-level F1 ($53.6$), ahead of the strongest evaluated LLM extractor ($48.7$).
\end{abstract}

\section{Introduction}

\label{sec:intro}
Researchers rely on scientific literature as a trusted source of information, but finding the relevant evidence in large paper collections remains difficult. Modern AI tools, especially those based on large language models (LLMs), offer a substantial increase in the efficiency of information search but introduce major risks to both individual users and organizations. Question Answering using LLMs lacks transparency and reliability. Even retrieval-augmented generation (RAG) systems \citep{Gupta:2024}, which use a combination of document retrieval and generative AI, are prone to most major issues of LLMs, including a tendency to produce factually inaccurate, irrelevant, or nonsensical output, commonly referred to as hallucinations \citep{Huang:2025}. Answers provided by LLMs cannot be trusted unless they are independently fact-checked, yet such verification remains a tedious process and is often omitted due to ``algorithm appreciation, where people tend to prefer algorithmic judgment over human judgment, even when the algorithm's processes are not fully transparent'' \citep{Logg:2019}

We present \textit{ACL-Verbatim}, an application of the VerbatimRAG framework for extractive question answering \citep{Kovacs:2025b} to the task of question answering from research papers in the ACL Anthology. While we contribute an end-to-end RAG system, complete with document preprocessing, indexing, and retrieval, our main focus is \textit{extraction}, the task of identifying spans in retrieved text chunks that are most useful for satisfying the information need conveyed by the user query. It is this step that differentiates VerbatimRAG from other RAG frameworks and which enables question answering without hallucinations.
%The overall architecture of our system is presented in Figure~\ref{fig:arch}.
In order to train and evaluate models on this task, we also create a pipeline for the automatic generation of search queries and perform manual annotation of text chunks to create a small ground truth dataset of 100 query-chunk pairs. Finally, we show that fine-tuning a compact extraction model on silver data generated by a strong LLM yields the best word-level F1 on our benchmark while using far fewer parameters than the evaluated LLM extractors.

%\begin{figure}[htbp]
%\caption{Overall architecture of the ACL-Verbatim system}
%\label{fig:arch}
%\end{figure}

\section{Related work}
\label{sec:related}

All question answering systems that allow an LLM to generate the final answer are prone to producing output that is factually incorrect, inconsistent with the provided evidence, or nonsensical, a phenomenon that is commonly referred to as hallucination \citep{Kaddour:2023,Huang:2025,Ji:2023}. It has therefore become generally accepted that answers provided by LLMs cannot be trusted for accuracy unless they are independently fact-checked, which defeats the purpose of applications in critical domains such as medical, legal, or financial question answering.

Retrieval Augmented Generation (RAG) has recently gained widespread popularity, but even though RAG systems reduce intrinsic hallucinations by grounding the model in external sources, extrinsic hallucinations can still occur due to LLMs' tendency to override retrieved information with their own prior ``knowledge''. RAG models continue to hallucinate \citep{Niu:2024}, limiting their use in complex and high-risk domains such as medical, legal, or financial question answering \citep{Lozano:2023,Magesh:2024}. A range of methods have recently been proposed for hallucination detection. Frameworks such as RAGAS \citep{Es:2024} and ARES \citep{SaadFalcon:2024} rely on specialized LLMs for large-scale hallucination detection but are not suitable for real-time prediction. Other LLM-based methods include approaches that use stochastic sampling \citep{Manakul:2023} or multi-step verification \citep{Friel:2023}. Classifier models trained on hallucination datasets such as RAGTruth \citep{Niu:2024} include RAG-HAT \citep{Song:2024}, RAGHalu \citep{Zimmerman:2024}, and LettuceDetect \citep{Kovacs:2025}. Approaches to hallucination detection that investigate connections between responses and source documents include Luna \citep{Belyi:2025} and FACTOID \citep{Rawte:2024}. Despite these recent efforts, hallucinations in RAG systems continue to limit the applicability of LLMs in real-world question answering tasks.

LLMs also suffer from lack of explainability, reducing both accountability and user trust. Mechanisms for generating post-hoc explanations of neural networks’ predictions are notoriously unreliable, and intuitive but wrong self-explanations offered by LLMs create additional risk by inflating users’ perception of their trustworthiness \citep{Madsen:2024,Chen:2024}. These risks are inherent to systems that allow neural language models to generate the final output presented to users, even when these models have been specialized for the domain of academic research \citep{Beltagy:2019,Taylor:2022,Viswanathan:2023}.

VerbatimRAG \citep{Kovacs:2025b} is an open-source RAG framework that tackles the issue of hallucinations by taking an extractive approach to retrieval-augmented question answering that only returns text spans that are taken verbatim from source documents. VerbatimRAG combines standard retrieval with extraction, the task of highlighting the parts of some input text that are relevant for answering some user query, for which the framework offers multiple approaches, including LLMs prompted for the extraction tasks as well as smaller models fine-tuned for the extraction/highlighting task, such as Provence \citep{Chirkova:2025} or the Zilliz Semantic Highlighter \citep{Zhang:2026}.

As generative models dominate most NLP applications, annotated benchmark datasets increasingly focus on abstractive rather than extractive approaches to question answering. Such vary in the source and genre of questions and answers, with a particular focus on general-domain knowledge using online sources such as Wikipedia and Reddit \citep{Stelmakh:2022,Fan:2019}, and many of them focus on \textit{factoid question answering}, where questions are expected to target specific facts that are present in some source and should be reproduced in the answer. ExpertQA \citep{Malaviya:2024} is a dataset that is also concerned with verification, containing expert annotations not only for system answers but also for the quality and reliability of cited sources.
The recent CLAPnq~\citep{Rosenthal:2025} dataset is of particular interest for the topic of extractive question answering.  Based on the Natural Questions benchmark \citep{Kwiatkowski:2019}, CLAPnq contains not only long-form answers to nearly 5k questions but also annotation of the subsets of sentences from retrieved passages that serve as the basis of these answers, making this dataset suitable for evaluating extractive models. The work closest to our application domain, which we also use as a basis for our query generation process to be described in Section~\ref{sec:annotation}, is SciRGen, a methodological framework for the large-scale generation of scientific QA datasets \citep{Lin:2025}.

Our experiments as well as our newly contributed dataset rely on the ACL Anthology\footnote{\url{https://aclanthology.org/}}, a public resource that has served as the basis of dozens of research datasets over the past decades \citep{Bollmann:2023}, including large-scale corpora such as NLP Scholar \citep{Mohammad:2020},  NLPExplorer\citep{Parmar:2020}, and the most recent ACL-OCL corpus \citep{Rohatgi:2023}, each of which provides valuable additional metadata for publications and enables advanced analysis of NLP research.

\section{Corpus creation}
\label{sec:corpus}

In this section we describe the corpus creation process, including data collection, preprocessing, segmentation, the generation of synthetic queries, as well as the human annotation process. Our pipeline is designed to allow for incremental updates of our dataset based on updates of the ACL anthology, detailed instructions are provided in the \texttt{acl-verbatim} repository\footnote{\url{https://github.com/KRLabsOrg/acl-verbatim}}. The version of the dataset that served as the basis for the annotation and evaluation described in this paper is based on the state of the ACL Anthology in February 2026.

\subsection{Data collection and preprocessing}
\label{sec:preproc}

\begin{figure*}[htb]
\includegraphics[width=\textwidth]{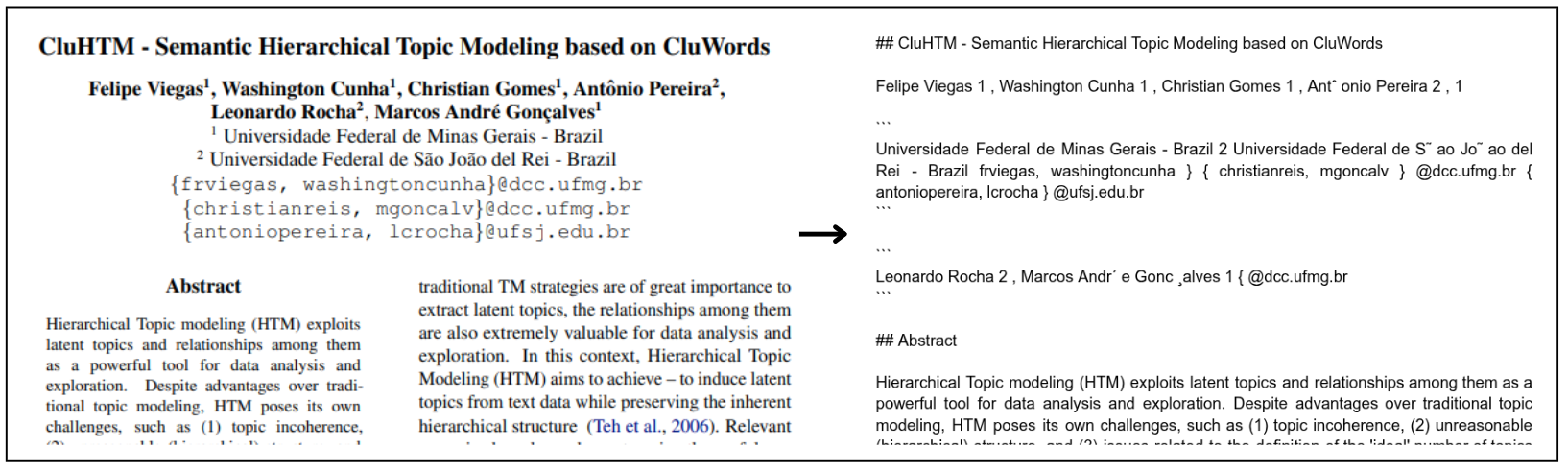}
\caption{Example of conversion from PDF to markdown using Docling}
\label{fig:docling_example}
\end{figure*}

The ACL Anthology \citep{Gildea:2018,Bollmann:2023} is a public library of over 120,000 research papers from the domains of computational linguistics and natural language processing. Metadata as well as full-text PDFs of papers are distributed under permissive licenses (CC BY 4.0 as of 2016) and programmatic access is provided via a GitHub repository\footnote{\url{https://github.com/acl-org/acl-anthology/}} and a Python library. We use these utilities to process all PDF files and to extract all paper metadata.  Downloading, filtering, and preprocessing of papers was based on metadata extracted from the ACL Anthology on February 26, 2026. Entries for 120 034 papers were processed, out of which 114 567 were mapped to PDFs for further processing. The remaining ~5k papers are hosted by third-party publishers and not covered by the permissive license of the ACL Anthology, these entries were discarded to maintain the flexible terms of the final dataset. 

PDFs are converted to markdown format using the open-source \texttt{docling} library\footnote{\url{https://docling-project.github.io/docling/}}, resulting in 114 475 markdown files (a total of fewer than 100 papers were skipped due to a variety of docling errors that were not further investigated). Docling's DocumentConverter is invoked using default settings. All text-based content including headers, lists, tables, figure captions etc. are rendered in markdown, while other figures and some formulas are replaced by placeholder text indicating that some content has been discarded. An example is shown in Figure~\ref{fig:docling_example}. We release this dataset of markdown files under a CC-BY 4.0 license on Huggingface\footnote{\url{https://huggingface.co/datasets/KRLabsOrg/acl-anthology-md}}.

Markdown documents are indexed using the VerbatimRAG library described in Section~\ref{sec:related}. For segmenting papers we implement a custom chunking strategy developed specifically for markdown-formatted research papers. This involves parsing section structure, segmenting papers along section boundaries, and prefixing each text chunk section and subsection titles to improve retrieval performance. The markdown chunker also prevents tables and code blocks from being split, and controls the minimum and maximum size of chunks, which we set to 500 and 5000 characters, respectively. Chunks are then indexed both for full text search and for dense vector search using the \texttt{granite-embedding-english-r2} embedding from IBM\footnote{\url{https://huggingface.co/ibm-granite/granite-embedding-english-r2}} \citep{Awasthy:2025}.

\subsection{Query generation and human annotation}
\label{sec:annotation}

This section describes the steps of creating a ground truth dataset mapping user queries to text spans in retrieved papers relevant for answering these queries. As a first step we create a sample of 333 papers in the ACL Anthology, randomly choosing from all English-language papers with at least one author (skipping full volumes that only have editors). Then we retrieve indexed chunks for these papers from the ACL-Verbatim index and randomly choose a single chunk for each paper. We then generate 3 synthetic queries for each chunk, following the ScIRGen methodology \citep{Lin:2025}. This two-step process involves prompting an LLM to generate a list of question types that could be answered by a given paragraph, then using in-context learning for each question type to generate questions. We extend this pipeline by a third step that converts long and linguistically sophisticated questions to shorter and more fragmented queries that are more characteristic of real-world user queries. An end-to-end example is presented in Figure~\ref{fig:query_gen_example}, full prompts are reproduced in Appendix~\ref{sec:prompts_q}.

\begin{figure*}[htb]
\includegraphics[width=\textwidth]{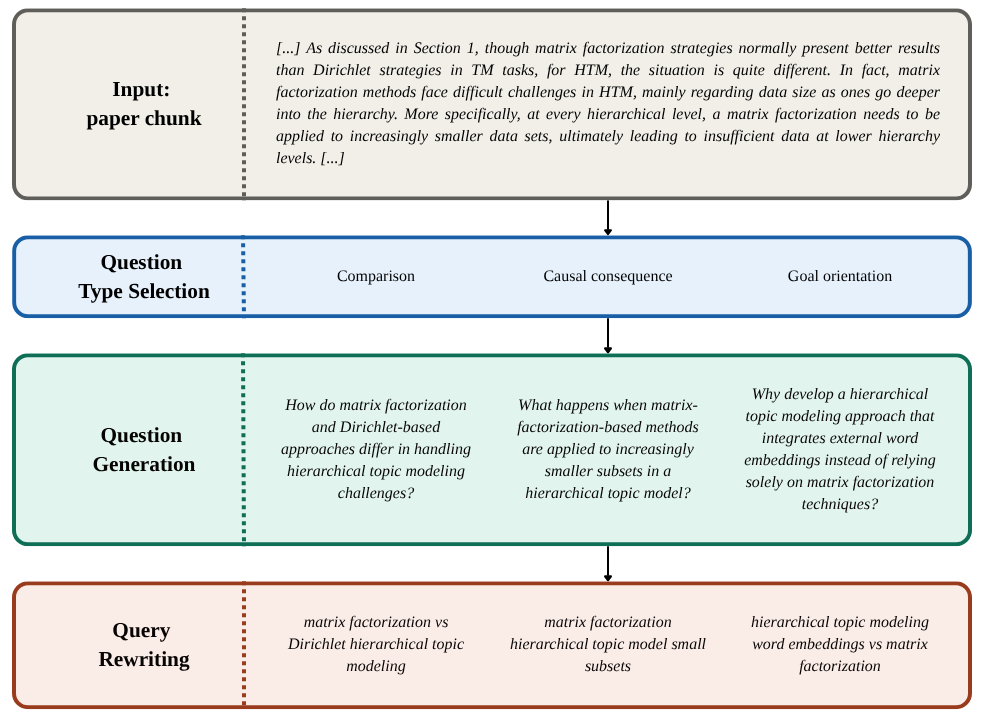}
\caption{Generation of synthetic user queries, based on the ScIRGen methodology. The example shows a chunk from the paper \textit{CluHTM - Semantic Hierarchical Topic Modeling based on CluWords} \citep{Viegas:2020}}
\label{fig:query_gen_example}
\end{figure*}

Generated queries are used to retrieve chunks from the VerbatimRAG index (see Section~\ref{sec:preproc}), and the top 5 chunks per query are used as input for the annotation task. Annotators then perform tasks for pairs of query and chunk. First, each chunk is annotated for relevance using a binary label. Chunks must be marked \texttt{relevant} if and only if the annotator considers the chunk to be relevant for satisfying the information need conveyed by the search query. If a chunk is labeled \texttt{irrelevant}, no further annotation takes place\footnote{Additionally, a small number of chunks were marked with a question mark to indicate that the relevance judgement doesn't make sense, this was the case in particular for chunks containing bibliography sections of papers. Such chunks were also not annotated further.}. For chunks considered relevant, annotators must also indicate (highlight) the span or spans of text within the chunk that are most relevant for answering the query. If a table or figure is considered relevant, annotators are instructed to highlight its caption. This way a sequential labeling task is defined, mapping each relevant text chunk to one or more continuous sequences of its tokens. Annotation is performed via an excel sheet, created programmatically from the output of the ACL-Verbatim system and postprocessed to create the JSON-formatted gold dataset. All components of the pipeline for query generation and annotation are published as open-source software on GitHub\footnote{\url{https://github.com/KRLabsOrg/acl-verbatim}}.

\subsection{Annotation challenges}

The core annotation task of determining the text fragments within a section of a research paper that are best suited to fulfill the information need conveyed by a search query is quite complex and raises several methodological issues. While the sampling process described above yielded a total of 906 queries, and retrieving the top 5 chunks for each produced 4530 query-chunk pairs, the human-annotated portion of our dataset contains only the first 20 queries and a total of 100 chunks. Annotation as well as the adjudication of differences among annotators was performed by authors of this paper, NLP researchers with some variety in their fields and level of experience. In this section we describe some of the key challenges we have encountered during the manual annotation process and discuss some implications for the extraction task underlying the \texttt{verbatim} approach to question answering. We illustrate issues with examples cherry-picked from the 20 queries in the manually annotated dataset.

A key challenge that is specific to the domain of academic research is that for many queries the annotation of the most relevant spans within a section of a paper requires considerable domain-specific expertise and careful consideration. For example, the (synthetic) query \textit{parsing merge predicate sequence equivalence conditions}, which is the simplified version of the more elaborate (synthetic) question \textit{What are the three conditions under which two instantiated sequences are considered equivalent by the parsing merge predicate?}, which in turn was generated based on a short section describing an algorithm in the paper \textit{LR Recursive Transition Networks for Earley and Tomita Parsing} \citep{Perlin:1991}. Retrieving the top 5 chunks from our index yields sections from 4 different papers. Perlin 1991 is not among them, but all of them are concerned with algorithms for syntactic parsing and all of them would appear to be potentially relevant based on their vocabulary. It is only by reading through and developing a basic understanding for each section that the annotators could make the first of two judgements, the binary decisions on whether these chunks are relevant at all and whether one should proceed with the extraction task to identify relevant spans. The final dataset categorizes two of the chunks as relevant, the introductory section of the 1989 EACL paper on \textit{Parsing and Derivational Equivalence} \citep{Hepple:1989} and a paragraph describing a core algorithm in the 2010 ACL paper \textit{Dynamic Programming for Linear-Time Incremental Parsing} \citep{Huang:2010}. Further annotation mapped both of these chunks to spans that should be extracted (highlighted) in response to the search query, which in one case reduced the 4700 character algorithm description to a single sentence of 92 characters (1.96\%) \textit{"The key observation for dynamic programming is to merge 'equivalent states' in the same beam"}, while the introductory section of the other paper was mostly relevant and only a few sentences were omitted from the extraction, reducing 1902 characters to 1447 (76.08\%). 

The above example is intended to illustrate the diversity of extracted spans, the difficulty and subjectivity of both annotation steps, and the meticulous effort required to create even a small high-quality dataset. Despite our efforts, one could argue that if our goal is to model the intent of the user hoping to find information by typing a search query \textit{parsing merge predicate sequence equivalence conditions}, then a reliable judgement on whether some retrieved section of a paper or any span of text within that section is a relevant result could only be made by an expert in the domain of parsing algorithms. While we consider this a valid argument, we relax this requirement in the interest of creating a novel and potentially useful dataset by allowing ourselves, NLP researchers with somewhat diverse backgrounds, to perform the annotation task, while also encouraging researchers to use our tools and data to create similar ground truth datasets for narrower domains as well as for use-cases other than academic research. 

%"## instance ~-  MERGE  (instance)\n\nASK instance\n\n- (1) Link up  with predecessor instances.\n- (2) Install self.\n- (3) ENQUEUE successor instances for insertion. }\n\nThe  parsing merge  predicate  considers  two instantiated sequences equivalent when:\n\n- (1) Their RTN  symbol  nodes  X are the same.\n- (2) They  are in  the same  column  k.\n- (3) They  have  identical left borders\n\ni.\n\nThe total number  of  links formed  by  INSERT  during an  entire parse, accounting for every grammar  RTN node, is O(n3)xO(IGI). The  chart parsers are a  family of  algorithms that couple efficient parse-tree merging with  various control organizations (Winograd, 1983).\n\n"

\section{Extraction experiments}
\label{sec:ext}

\subsection{Extraction models}

We evaluate extractive models on the manually annotated benchmark introduced in the previous sections. The benchmark contains 20 synthetic queries paired with the top-5 retrieved chunks per query, yielding 100 query--chunk pairs in total. Of these, 47 chunks are annotated as relevant and contain 78 gold evidence spans, while the remaining 53 chunks are irrelevant and have no gold spans. We report extraction metrics on all 100 rows.

Three families of extractive systems are compared. First, we evaluate LLM-based span extractors. These models receive a question together with the retrieved chunk and must return verbatim evidence spans. We evaluate Mistral Small 2603, Nemotron-120B-A12B, GLM-5, and Qwen 3.6 35B. For Mistral, Nemotron, and Qwen, we compare a default extraction prompt against a paragraph-oriented prompt designed to encourage broader evidence selection.
Second, we evaluate extractive pruning and highlighting baselines. Zilliz Semantic Highlight \citep{Zhang:2026} selects relevant sentences or token spans from the chunk, while Provence \citep{Chirkova:2025} prunes irrelevant sentences from the context using a DeBERTa-v3 reranker-style architecture. Zilliz follows the same general token-scoring formulation as Provence, but is trained as a bilingual semantic-highlighting model on top of a BGE-M3 reranker backbone \citep{ChenJianlv:2024}. Provence operates with a native context budget of 512 tokens and internally splits longer chunks. Finally, we also train a compact student model on silver supervision, which we describe in Section~\ref{sec:supervision}.

\subsection{Supervision}
\label{sec:supervision}

To train a self-contained student model, we generate silver supervision from the ACL Anthology corpus using synthetic queries and retrieved chunks, following the steps described in Section~\ref{sec:annotation}. The current release is based on 2000 sampled papers, which yielded 5892 synthetic queries, 32480 raw silver query--chunk rows, and 23235 retained rows after filtering. The final split contains 20916 silver training rows and 2319 development rows in canonical form, and 20920 / 2319 tokenized training / development windows at the full 8192-token ModernBERT context. At this context length roughly every silver row fits in a single window; the total amount of silver supervision the student sees is approximately 10k positive rows with spans and 11k negative rows. Evidence density (span characters over chunk characters) is $11.7\%$, i.e. a roughly $1{:}8$ token-level class imbalance.

Our student architecture is a query-conditioned token classifier over an 8192-token ModernBERT backbone, with binary token labels and sliding-window inference. The input is the concatenation of
question and chunk, and the output is a binary evidence label per token, decoded into character spans. We compare two backbones: the vanilla \texttt{answerdotai/ModernBERT-base} MLM checkpoint and the
\texttt{Alibaba-NLP/gte-reranker-modernbert-base} cross-encoder, which has been post-trained on query--passage relevance.
The silver teacher is Qwen 3.6 35B with the paragraph-oriented extraction prompt. We train for 5 epochs with batch size 8 at learning rate $2{\times}10^{-5}$; the best checkpoint is selected by silver-dev token F1. Two post-processing steps are applied at inference: spans shorter than 10 characters are dropped, and neighbouring spans separated by at most 20 characters are merged. These two rules remove token-level fragmentation in which the model emits a "shotgun" of short pseudo-spans around genuine evidence tokens. The student model based on the reranker backbone is released as \texttt{KRLabsOrg/acl-verbatim-modernbert}.

\subsection{Evaluating extraction}
\label{sec:metrics}

We propose several overlap-based metrics for comparing extracted text spans against the ground truth on all 100 query--chunk rows in the benchmark. Our primary method of evaluation and comparison is word-level precision and recall, which compares the sets of words covered by gold and extracted spans. We prefer this metric because it isolates a model's ability to highlight the right words in a piece of text and is not sensitive to whether span boundaries are correctly predicted and whether models (and annotators) prefer fewer and longer spans or many shorter ones. The best configurations for each model type were selected by comparing word-level F1 scores. Two additional, asymmetric measures are used as alternative approaches to comparing pairs of span sets. \textit{Containment} measures whether a large enough part of predicted spans are contained by gold spans. \textit{Containment @ 1} is the ratio of predicted spans that are fully contained by a gold span, \textit{Containment @ 0.8} is the ratio of spans that are at least 80\% covered by a gold span, etc. Analogously, \textit{Coverage} measures the ratio of gold spans that are covered by predicted spans to some degree, e.g. \textit{Coverage @ 0.8} is the number of gold spans that are at least 80\% covered by predicted spans, divided by the total number of gold spans. Overlaps between spans are measured at the character level.

The right choice of metrics for evaluating span extraction depends heavily on the nature of the extraction task. One of the most common span extraction tasks in natural language processing is Named Entity Recognition (NER), where system outputs are commonly evaluated using span-level precision and recall, which requires exact matches between span boundaries and will penalize even the smallest mismatch by considering it as both a false positive and a false negative. We argue for our choice of metrics with a simple example. Consider a sequence of 100 characters in a retrieved chunk of text that is annotated as containing two relevant spans of 45 characters each, with a 10 character long break between them. Then consider a system that predicts this entire span of 100 characters as relevant, i.e. its mistake is merging the 10 irrelevant characters with the 90 relevant ones. This prediction would achieve span-level precision and recall scores of zero, since it has not made any correct predictions. In contrast, the word-level precision and recall  of the system are $0.9$ and $1.0$, respectively. Finally, the \textit{Containment @ t} ratio is 1 for all $t<0.9$ and 0 for $t\geq0.9$, while \textit{Coverage @ t} is 1 for any value of $t$, expressing alternative preferences in evaluating extraction.

In addition to the ACL-specialized model evaluated here, we release a multi-domain sibling model, \texttt{KRLabsOrg/verbatim-rag-modern-bert-v2}\footnote{\url{https://huggingface.co/KRLabsOrg/verbatim-rag-modern-bert-v2}}, trained on \texttt{KRLabsOrg/verbatim-spans}\footnote{\url{https://huggingface.co/datasets/KRLabsOrg/verbatim-spans}}. This dataset combines our ACL silver data with RAGBench \citep{Friel:2025}, a large-scale benchmark of retrieval-augmented question answering examples across industry-oriented domains, and Squeez \citep{Kovacs:2026}, a task-conditioned tool-output pruning dataset built from coding-agent tool observations. We evaluate this generic model separately in its model card, including QASPER \citep{Dasigi:2021}, a scientific QA benchmark over NLP papers, as an out-of-training-domain test set. On ACL gold the generic model reaches $0.463$ word-level F1, compared to $0.301$ for Zilliz Semantic Highlight and $0.344$ for Provence, and it also outperforms both baselines on RAGBench, Squeez, and QASPER.

\section{Results}

\begin{table*}[htb]
  \centering
  \small
  \begin{tabular}{lrrrrrrrrrr}
  \toprule
  Model & \multicolumn{3}{c}{\textit{Word}} & \multicolumn{3}{c}{\textit{Containment @ t}} & \multicolumn{3}{c}{\textit{Coverage @ t}}& Latency \\
  \cmidrule(lr){2-4} \cmidrule(lr){5-7} \cmidrule(lr){8-10}
  & Prec. & Rec. & F1 & 1.0 & 0.8 & 0.5 & 1.0 & 0.8 & 0.5 & \\
  \midrule
  \multicolumn{11}{l}{\textit{LLM extractors}} \\
  Mistral Small 2603\textsuperscript{a}              & 40.41 & 55.99 & 46.94 & 43.86 & 46.24 & 49.73 & 17.95 & 39.74 & 57.69 & 0.78 \\
  Mistral Small 2603\textsuperscript{b}  & 34.22 & \textbf{73.03} & 46.61 & 27.70 & 33.61 & 39.04 & 48.72 & \textbf{65.38} & \textbf{75.64} & 1.07 \\
  Nemotron-120B-A12B\textsuperscript{a}              & 34.14 & 50.92 & 40.88 & 36.62 & 36.62 & 42.88 & 43.59 & 43.59 & 52.56 & 0.38 \\
  Nemotron-120B-A12B\textsuperscript{b}  & 30.15 & 62.72 & 40.73 & 29.93 & 31.29 & 44.90 & \textbf{58.97} & 62.82 & 62.82 & 0.54 \\
  GLM-5\textsuperscript{a}                           & 44.50 & 53.80 & 48.71 & \textbf{52.87} & \textbf{54.25} & \textbf{58.36} & 20.51 & 34.62 & 53.85 & 1.04 \\
  Qwen 3.6 35B\textsuperscript{a}                    & 44.74 & 40.88 & 42.72 & 49.05 & 50.95 & 53.33 & 15.38 & 25.64 & 41.03 & 1.63 \\
  Qwen 3.6 35B\textsuperscript{b}         & 39.43 & 57.35 & 46.73 & 34.78 & 36.96 & 47.83 & 44.87 & 51.28 & 61.54 & 1.20 \\
  \midrule
  \multicolumn{11}{l}{\textit{Pruning / highlighting baselines}} \\
  Zilliz Semantic Highlight\textsuperscript{c}        & 46.97 & 22.11 & 30.07 & 33.33 & 35.46 & 39.53 & 2.56 & 8.97 & 19.23 & 1.04 \\
  Provence\textsuperscript{d}                         & 27.58 & 45.70 & 34.40 & 21.56 & 22.84 & 29.23 & 16.67 & 21.79 & 41.03 & 2.40 \\
  \midrule
  \multicolumn{11}{l}{\textit{Our students (150M params)}} \\
 acl-verbatim-modernbert \textsuperscript{e}              & \textbf{65.43} & 45.43 & \textbf{53.63} & 27.27 & 31.47 & 48.22 & 35.90 & 37.18 & 42.31 & 0.47 \\
   verbatim-rag-modern-bert-v2 \textsuperscript{f}               & 63.00 & 36.58 & 46.29 & 30.30 & 31.82 & 46.97 & 26.92 & 32.05 & 38.46 & 0.40 \\
  \bottomrule
  \end{tabular}
  \caption{Extractor results on the gold benchmark (100 query--chunk pairs: 47 relevant and 53 irrelevant). See Section~\ref{sec:metrics} for metric definitions. Values are percentages, except latency, which is seconds per row. Latencies for Zilliz, Provence, and ModernBERT were measured on CPU.}
  \label{tab:extractor_results}

  \vspace{2pt}
  \raggedright
  \footnotesize
  \textsuperscript{a}\,Default extraction prompt.
  \textsuperscript{b}\,Paragraph-oriented extraction prompt.
  \textsuperscript{c}\,Zilliz token-span output at threshold 0.3.
  \textsuperscript{d}\,Native Provence setting with internal splitting for contexts longer than 512 tokens.
  \textsuperscript{e}\,Threshold 0.2 with min-span length 10 and merge gap 20.
  \textsuperscript{f}\,\texttt{KRLabsOrg/verbatim-rag-modern-bert-v2}, threshold 0.2 with min-span length 30 and merge gap 20.
\end{table*}

Table~\ref{tab:extractor_results} compares extractor models on the 100 rows of the manually annotated benchmark. The best Word-F1 is achieved by our reranker-initialized ModernBERT student ($53.63$), ahead of the strongest LLM extractors, GLM-5 ($48.71$), Mistral Small ($46.94$), and Qwen with the paragraph prompt ($46.73$), while using 3 to 4 orders of magnitude fewer parameters. The generic multi-domain ModernBERT model also remains competitive on ACL gold ($46.29$ Word-F1), outperforming the public extractive baselines despite not being specialized only for the ACL Anthology. Our ACL-specialized model also achieves the highest word-level precision. Unlike the LLM extractors, it often abstains on irrelevant chunks. LLMs, in particular those used with paragraph-oriented prompts, achieve higher recall and higher span-level coverage, but achieve substantially lower precision, extracting evidence from many chunks that are irrelevant for the query. This trade-off is of particular importance in our context of retrieval-augmented question answering, where high-precision extraction models are effective filters of irrelevant search results.  This difference is directly observable if we compare the results of our best model with an LLM-based extractor that achieves much higher recall. On the 100 chunks in the evaluation dataset, 53 of which had no gold spans annotated, our model chose not to predict any spans for 60 chunks while the paragraph-based Mistral model abstained only 35 times. We also illustrate this behavior with a cherry-picked example. For the query \textit{hate speech detection downsampled training examples number}, one of the top retrieved chunks is a subsection describing the experimental dataset in the paper \textit{EDAL: Entropy based Dynamic Attention Loss for HateSpeech
Classification} \citep{Fahim:2023}. This text provides lots of detail about topics closely related to the query, including statistics on class labels, but it is not at all concerned with downsampling. Our model correctly chose not to extract any spans, and so did the Zilliz highlighting model and some of the LLM-based extractors, including the default Mistral model, both Nemotron models, and the paragraph-based Qwen model. However, the four remaining models each choose some false positive spans, including texts on merging labels and texts as well as tables on dataset sizes.

%Provence is the stronger pruning baseline by recall, while Zilliz is more conservative and has higher word precision but much lower recall.

\section{Conclusion}
We described an application of the VerbatimRAG architecture to over 100K research papers in the ACL Anthology, contributed a manually annotated dataset for the core extraction task, and presented a set of experiments showing that small customized encoder-decoder architecture trained with synthetic data outperforms zero-shot LLM-based extraction on this task, at a fraction of the cost. We release all components of our pipeline as open-source software. We believe that combining the VerbatimRAG approach with the task-oriented training of extractive models provides a blueprint for the efficient deployment of high-performing hallucination-free question answering systems across a variety of domains.

\section*{Limitations}
The validity of our conclusions is limited by the size of the manually annotated dataset that was the basis of both quantitative and qualitative evaluation. The high complexity of the annotation task, described in detail in Section~\ref{sec:annotation}, also limited our ability to measure agreement between multiple annotators. to implement a rigorous adjudication process for resolving differences among annotators, or to develop detailed and objective annotation guidelines. We believe that all these steps will be possible if our approach is applied to more narrowly defined question answering use-cases that in turn lead to more objective extraction tasks. Furthermore, the extraction models trained using synthetic training data may reproduce unintended bias present in the output of LLMs, which may lead to such bias being reinforced and propagated by our models.

\section*{Acknowledgments}
GR implemented the pipelines for data processing and query generation, performed manual annotation, and implemented parts of the evaluation. ÁK designed and executed extraction experiments. NV participated in the annotation and contributed to literature research. SzT and IB participated in the implementation and execution of retrieval and extraction experiments. 
Work partially supported by the ``CLEAR" project, funded within the Cybersecurity Programme Kybernet-Pass of the Austrian Federal Ministry of Finance and managed by the Austrian Research Promotion Agency.

\bibliography{acl_verbatim}

 \appendix

\section{Detailed model comparison}
\label{sec:appendix_full_metrics}

Table~\ref{tab:appendix_full} reports containment and coverage metrics for the extractor configurations evaluated in Table~\ref{tab:extractor_results}. All metrics are computed on the full 100-row benchmark, including the 53 irrelevant query--chunk pairs as negative examples.

\begin{table*}[htbp]
\centering
\small
\begin{tabular}{lrrrrrr}
\toprule
 & \multicolumn{3}{c}{Containment @ t} & \multicolumn{3}{c}{Coverage @ t} \\
\cmidrule(lr){2-4} \cmidrule(lr){5-7}
Model & 0.5 & 0.8 & 1.0 & 0.5 & 0.8 & 1.0 \\
\midrule
Mistral Small 2603 & 0.497 & 0.462 & 0.439 & 0.577 & 0.397 & 0.179 \\
Mistral Small 2603 + paragraph & 0.390 & 0.336 & 0.277 & 0.756 & 0.654 & 0.487 \\
Nemotron-120B-A12B & 0.429 & 0.366 & 0.366 & 0.526 & 0.436 & 0.436 \\
Nemotron-120B-A12B + paragraph & 0.449 & 0.313 & 0.299 & 0.628 & 0.628 & 0.590 \\
GLM-5 & 0.584 & 0.542 & 0.529 & 0.538 & 0.346 & 0.205 \\
Qwen 3.6 35B & 0.533 & 0.510 & 0.490 & 0.410 & 0.256 & 0.154 \\
Qwen 3.6 35B + paragraph & 0.478 & 0.370 & 0.348 & 0.615 & 0.513 & 0.449 \\
Zilliz Semantic Highlight & 0.395 & 0.355 & 0.333 & 0.192 & 0.090 & 0.026 \\
Provence reranker-pruner & 0.292 & 0.228 & 0.216 & 0.410 & 0.218 & 0.167 \\
ACL-Verbatim GTE-reranker ($t{=}0.2$ + merge) & 0.482 & 0.315 & 0.273 & 0.423 & 0.372 & 0.359 \\
\bottomrule
\end{tabular}
\caption{Detailed extractor metrics on the full 100-row gold benchmark. Containment measures how much of a predicted span lies inside a gold span; coverage measures how much of a gold span is covered by a prediction.}
\label{tab:appendix_full}
\end{table*}

\section{Threshold selection for the student model}
\label{sec:appendix_threshold}

The student is a binary token classifier, so span decisions depend on a probability threshold at inference. Table~\ref{tab:appendix_threshold} reports the all-row gold benchmark scores for the GTE-reranker student at $t \in \{0.2, 0.3, 0.4, 0.5\}$ with the same post-processing held constant: spans shorter than 10 characters are dropped, and neighbouring spans separated by at most 20 characters are merged.

The best Word-F1 is obtained at $t{=}0.2$. Increasing the threshold improves precision but reduces recall, which lowers F1 on this benchmark.

\begin{table}[t]
\centering
\small
\begin{tabular}{lrrrr}
\toprule
Threshold & Word-P & Word-R & Word-F1 & IoU-F1@0.5 \\
\midrule
0.2 & 0.654 & 0.454 & \textbf{0.536} & \textbf{0.389} \\
0.3 & 0.667 & 0.421 & 0.516 & 0.338 \\
0.4 & 0.678 & 0.403 & 0.506 & 0.365 \\
0.5 & \textbf{0.701} & 0.380 & 0.493 & 0.336 \\
\bottomrule
\end{tabular}
\caption{Threshold sweep for the GTE-reranker student on the full 100-row gold benchmark.}
\label{tab:appendix_threshold}
\end{table}

\section{Prompts for query generation}
\label{sec:prompts_q}

  \subsection{Question-type classification prompt}

  \begin{lstlisting}
  You are a researcher generating questions and answers to find relevant
  information within a specific domain. Below are the potential question types.
  Choose the type that best fits the field information and the user's purpose.

  1. Verification: questions seeking a simple yes/no confirmation.
  2. Disjunctive: questions presenting multiple alternatives.
  3. Concept Completion: questions starting with Who/What/When/Where.
  4. Example: questions asking for instances of a concept.
  5. Feature Specification: questions about properties or characteristics.
  6. Quantification: questions seeking numerical or measurable information.
  7. Definition: questions asking for the meaning of a term or concept.
  8. Comparison: questions asking for similarities or differences.
  9. Interpretation: questions asking for inference over observed patterns.
  10. Causal Antecedent: questions about causes or reasons.
  11. Causal Consequence: questions about outcomes or results.
  12. Goal Orientation: questions about objectives or intentions.
  13. Instrumental/Procedural: questions asking how to achieve a goal.
  14. Enablement: questions about conditions enabling an action.
  15. Expectation: questions about anticipated or missing outcomes.
  16. Judgmental: questions asking for evaluation or opinion.
  17. Assertion: statements indicating lack of knowledge.
  18. Request/Directive: requests to summarize, analyze, or search.

  Task: Based on the following text from a research paper, return the most
  appropriate 3 question types that could be answered by this text. Give me the
  name of each type and not other information. Return ONLY valid JSON -- an
  array of objects, no markdown or explanations.

  Text: {chunk}
  \end{lstlisting}

  \subsection{Question generation prompt}

  \begin{lstlisting}
  You are a researcher asking questions aiming to find information in research
  papers.

  Content of paper: {chunk}

  Please generate one question that can be answered by the above text and which
  belongs to the question type below.

  - Question Type: {q_type}
  - Question Description: {q_def}
  - Question Example: {q_ex}

  Instructions:
  1. Only return a question without any other information.
  2. Use neutral terms like "a dataset", "data collection method", or
     "research approach", instead of references like "the study" or
     "this dataset".
  3. The question should be short and simple, resembling what a user might
     type into a search engine.
  4. The question should be answerable based on the text above.
  \end{lstlisting}

  \subsection{Query rewriting prompt}

  \begin{lstlisting}
  You are a researcher using a search engine to find information.

  Your question: {question}

  Please generate a search query that you would use to find the answer to this
  question.

  Instructions:
  1. Only return a search query without any other information.
  2. The query should be short and simple, resembling what a user might type
     into a search engine.
  3. The query does not need to be grammatical.
  \end{lstlisting}

\section{Prompts for extraction}
\label{sec:prompts_ext}

  \subsection{Default VerbatimRAG extraction prompt}

  \begin{lstlisting}
  Extract EXACT verbatim text spans from multiple documents that answer the
  question.

  Rules
  1. Extract only text that explicitly addresses the question.
  2. Never paraphrase, modify, or add to the original text.
  3. Preserve original wording, capitalization, and punctuation.
  4. Order spans within each document by relevance, most relevant first.
  5. Include complete sentences or paragraphs for context.

  Output format
  Return a JSON object mapping document IDs to span arrays ordered by relevance:
  {
    "doc_0": ["most relevant span", "next most relevant span"],
    "doc_1": ["most relevant from doc 1"],
    "doc_2": []
  }

  If no relevant information exists in a document, use an empty array.

  Your task
  Question: {{ question }}

  Documents:
  {{ documents }}

  Extract verbatim spans from each document:
  \end{lstlisting}

  \subsection{Paragraph-style extraction prompt}

  \begin{lstlisting}
  Extract verbatim supporting passages from each document that answer the
  question.

  What to extract
  A supporting passage is the complete portion of the document a researcher
  would highlight to justify the answer, including:
  - the sentence(s) that directly address the question;
  - preceding setup sentence(s) that introduce the topic, methodology, or
    figure being referenced;
  - concluding interpretation sentence(s) that summarize implications;
  - table captions when the table itself is relevant.

  Prefer a single continuous paragraph over multiple fragments of the same
  paragraph. Only split into multiple spans when relevant content is in
  non-adjacent parts of the document.

  Rules
  1. Use EXACT text from the document; no paraphrasing or edits.
  2. Preserve original wording, capitalization, and punctuation.
  3. If no passage in the document supports the answer, return an empty array.
  4. Order spans within each document by relevance, most relevant first.

  Output format
  Return JSON mapping document IDs to arrays:
  {
    "doc_0": ["first supporting passage", "second supporting passage"],
    "doc_1": ["passage from doc 1"],
    "doc_2": []
  }

  Your task
  Question: {{ question }}

  Documents:
  {{ documents }}

  Extract supporting passages from each document:
  \end{lstlisting}
\end{document}